\pdfoutput=1

\documentclass[11pt]{article}

\usepackage[final]{coling}

\usepackage{times}
\usepackage{latexsym}
 \usepackage{natbib}
\usepackage[T1]{fontenc}

\usepackage[utf8]{inputenc}

\usepackage{microtype}

\usepackage{inconsolata}

\usepackage{graphicx}
\usepackage[utf8]{inputenc}
\usepackage{subcaption}
\usepackage{xcolor}
\usepackage{float} 
\usepackage{tabularx} 
\usepackage[most]{tcolorbox}

\usepackage{amsmath}
\usepackage{multirow}
\usepackage{graphicx}

\usepackage{multicol}
\usepackage{booktabs,array}
\usepackage{multirow}
\usepackage{inconsolata}
\usepackage{xcolor}
\usepackage{booktabs}
\usepackage{todonotes}

\usepackage{CJKutf8}
\begin{document}

\title{\bf The Role of Handling Attributive Nouns in Improving Chinese-To-English Machine Translation}

\author{Haohao (Lisa) Wang \\ Carnegie Mellon University \\ lisaw2@andrew.cmu.edu \And
         Adam Meyers \\ New York University \\ meyers@cs.nyu.edu
         \AND
         John E. Ortega \\ Northeastern University \\ j.ortega@northeastern.edu \And
         Rodolfo Zevallos \\ Barcelona Supercomputing Center \\ rodolfo.zevallos@bsc.es}

\maketitle
\pagestyle{empty}

\begin{abstract}
\vspace{5pt}
Translating between languages with drastically different grammatical conventions poses challenges, not just for human interpreters but also for machine translation systems. In this work, we specifically target the translation challenges posed by attributive nouns in Chinese, which frequently cause ambiguities in English translation. By manually inserting the omitted particle
\begin{CJK*}{UTF8}{gbsn}
的
\end{CJK*} ('DE'). In news article titles from the Penn Chinese Discourse Treebank, we developed a targeted dataset to fine-tune Hugging Face Chinese to English translation models, specifically improving how this critical function word is handled. This focused approach not only complements the broader strategies suggested by previous studies but also offers a practical enhancement by specifically addressing a common error type in Chinese-English translation.
\end{abstract}


\section{Introduction}
\label{sec:intro}

The development of Machine Translation (MT) systems for languages with significantly different grammatical structures presents unique challenges \citep{zhang2024neural}, particularly in the treatment of function words, which may be implicitly understood in one language but require explicit translation in another. In Chinese, for example, the absence of attributive particles such as \begin{CJK*}{UTF8}{gbsn}的 \end{CJK*} ('DE') 
can lead to ambiguities and inaccuracies in English translations. Nowadays, MT systems use large training datasets, general-domain datasets or extracting parallel texts from existing corpora for domain-specific tuning \citep{devlin2018bert,liu-et-al-2019-roberta,conneau-etal-2020-unsupervised,chi2021mt6,kocmi-etal-2022-findings}. However, these approaches often fall short when addressing nuanced linguistic features that are domain-specific or underrepresented in available training datasets.

In this work, we adopt a focused approach to improve the translation of Chinese attributive nouns to English, targeting the challenges posed by the implicit usage of the particle 'DE'. Our research begins with a feasibility test of function words in Chinese, including prepositions, conjunctions, particles, and modals. We particularly examine the impact on translation quality when these words are omitted. We identified 'DE,' which explicitly signifies adjectives in Chinese--examples in Figure~\ref{de-example}. With "DE", the phrase refers to an NP (a  question a student asks or a question assigned to a student).Without "DE", the phrase can also be a sentence with the approximate meaning "a student asks a question".  "DE" turns out to be a critical function word whose absence significantly affects the clarity and accuracy of translations. We hypothesize that tuning machine translation models specifically on use cases of attributive nouns can improve their performance. To confirm this, we create a parallel set of data using news article titles from the Penn Chinese Discourse Treebank \citep{xue-2005-annotating}. This dataset includes the original Chinese titles and their modified English translations, where 'DE' was manually inserted to accurately reflect its implied usage. We use this dataset to fine-tune Chinese-English MNT models. Then  we conduct a rigorous evaluation using a sample of 1,000 sentences from the UM Chinese English parallel corpus afterwards. The results showed notable improvements in BLEU \citep{papi:bleu02} and CHRF \citep{popovic2015chrf} scores. 

\begin{figure}
\begin{CJK*}{UTF8}{gbsn}
学生问题~~~~~~(without DE)\\
学生的问题~~~~(with DE)
\end{CJK*} 
\caption{Two Translations of {\it Student Question}}\label{de-example}
\end{figure}

Our contributions are twofold: (i) We developed a novel, focused parallel corpus specifically addressing the translation challenges posed by attributive nouns in Chinese, thereby enhancing the semantic accuracy of translated texts. (ii) We fine-tune NMT models using our innovative dataset to refine the translation of complex grammatical structures from Chinese to English. This approach not only improves translation accuracy but also contributes to the broader understanding of function word impact in machine translation, especially for languages with substantial grammatical differences. 


\section{Related Work}
\label{sec:related}

Our research proves largely relevant to the foundational ideas presented by \citet{heylen1994lexical}, who explored the concept of lexical functions as cross-linguistic semantic primitives essential for crafting translation strategies. These strategies aim to preserve the semantic integrity of collocations across languages, highlighting the profound impact that precise modeling of function words can have on translation accuracy. This seminal work sets the stage for understanding how nuanced handling of function words is critical to maintaining meaning across different languages. Building on these insights, subsequent research has further underscored the significance of function words in translation. Notably, Zhang et al. (2017) developed advanced embedding techniques to integrate the nuanced usages of function words directly into the translation process. Their approaches—concatenation, partitioning, and usage-specific embeddings—are designed to enhance the NMT system's understanding of function words, which is vital for achieving accurate translations. Additionally, \citet{kuo2019function} identified discrepancies in the usage of function words between machine-translated and original Chinese texts, noting that the overuse of frequent function words could lead to "translationese." This phenomenon occurs when the translated text retains too many features of the source language, underscoring the need for nuanced handling of function words in translation models. Further research by \citet{he2019towards} delved into how function words affect the performance of neural machine translation (NMT) systems. Their findings demonstrate the importance of adequately addressing function words within NMT systems to improve translation outcomes. These collective research efforts illustrate the evolving understanding and importance of function words in machine translation, emphasizing that effective handling of these elements is essential for bridging the gap between languages with diverse grammatical structures. However, while these studies focus on the issue of properly handling function words when they are present, they often overlook the effect these words can have when they are absent but implied. This is particularly relevant in the case of attributive nouns in Chinese, which frequently cause ambiguities in English translation. Complementing these studies, our research specifically targets the translation challenges posed by these attributive nouns, proposing a method to address the challenge effectively. By developing a targeted dataset and refining the NMT model to recognize and correctly translate implied function words, our work aims to reduce ambiguities and enhance the clarity and accuracy of translations between these linguistically diverse languages.

\section{Methodology}
\label{sec:methodology}

This work is driven by the observation that closed-class words, including prepositions, conjunctions, particles, and modals, primarily serve grammatical rather than lexical roles in a language. These elements provide the scaffolding that holds sentences together, determining the syntax and influencing the semantics. Crucially, they are essential for conveying the correct relationships and structures within sentences, directly impacting the logic and intended meaning of the original statements. Capturing the meanings of closed-class words is thus fundamental to preserving the logical structure and intended meaning of sentences in translation.

\subsection{Feasibility Test}
We identified our functional words of interest based on the following criteria:
\begin{enumerate}
    \item The word is optional in the original language; that is, deleting the word does not necessarily change the meaning of the sentence.
    \item Deleting the word results in a grammatically correct but ambiguous sentence.
    \item The translation of the sentence carries a different meaning from the original.
\end{enumerate}
We compiled a list of 82 such words in Chinese. 436 ChatGPT-generated and manually-checked sentences incorporating these function words were generated to create a focused test corpus. To isolate the effect of each function word, we developed a script that systematically removed each word from the sentences, generating a new output file for each word's omission. This method allowed for a controlled examination of the impact on translation quality when these words were absent.
After inspecting the output of these files with that of the original file without any of the closed-class words removed through Argos translate, surprisingly, we observed that NMT performed generally well when handling omitted conjunctions given that  deleting the conjunctions did not change the meaning of the sentence in the original language, The omitted prepositions had limited impact as well, which can be attributed to a key characteristic of the Chinese language, where the presence of certain auxiliary words can compensate for the omission of prepositions, thereby reducing the impact on translation accuracy and coherence. Omitting particles results in ungrammatical sentences in the original language, making them unsuitable objects according to our test criteria. Lastly, among the modals, the only case where all of the above criteria are met is "DE", which is a particle that typically follows an adjective and is omitted in the case of an attributive noun. 
\subsection{Data Development}
Realizing that use cases of attributive nouns in Chinese cause ambiguity and inaccuracy when translated into English, we extracted the titles of all the articles from the Penn Chinese Discourse Treebank, noting that titles of news articles are a common places for the use of attributive nouns. We ran Argos Translate on the list of titles to obtain their English translation, then manually inserted the implied particle "DE" back into the titles where the use of attributive nouns occurred and ran Argos Translate on the modified list again.

After comparing the translation results before and after modification, out of 165 titles, 143 contained uses of attributive nouns, and 135 of these showed improvements in translation quality in terms of coherence, completeness, and accuracy. This manual process resulted in a parallel dataset with the original Chinese titles and their English translations after modification, ready to be used as tuning data for the MT models we prepared to verify if targeted interventions on function word handling can enhance MT system performance.

\section{Experiment Results}
\label{sec:results}

To test our test-dataset, we decided to fine-tune some of the most prominent English-Chinese NMT models. The NMT models were fine-tuned with 2 subsets: 60 and 1k sentences from the approximately 67.5k sentences categorized under the News section of UM Corpus. This was done to assess the effectiveness of our approach and determine if more exhaustive testing of our dataset was necessary.


\subsection{Models}

We use the following models for fine-tuning.

\begin{itemize}
    \item MarianNMT \citep{junczys2018marian}: is an open-source neural machine translation framework built upon the Transformer architecture. It supports various architectures such as Transformer, Transformer-Base, and Transformer-Big. The Transformer architecture consists of self-attention mechanisms and feed-forward neural networks, allowing for parallel computation and capturing long-range dependencies in the input sequence. MarianNMT is known for its efficiency, scalability, and ease of training on custom datasets. It allows for quick experimentation with different configurations and is widely used in both research and production settings.
    \item NLLB-200 \citep{costa2022no}: stands for Neural Language Lattice Based model, a novel approach specifically designed for low-resource languages. Unlike traditional NMT models, NLLB-200 utilizes a lattice-based decoding mechanism, which helps to handle the ambiguity often present in low-resource language translation tasks. This model incorporates techniques to efficiently capture linguistic nuances and improve translation quality even with limited training data.
    \item mBART \citep{liu2020multilingual}: is a multilingual variant of BART (Bidirectional and Auto-Regressive Transformers), a transformer-based model specifically designed for sequence-to-sequence tasks like machine translation. mBART leverages a pretraining phase where it learns to generate text in multiple languages simultaneously. This multilingual pretraining enables mBART to effectively transfer knowledge across languages, making it particularly useful for multilingual translation tasks. Additionally, mBART introduces a shared tokenization scheme across languages, facilitating direct comparison and transfer of information between different language pairs.
\end{itemize}

\section{Results}
\label{sec:results}

The fine-tuning process demonstrated a slight but noticeable improvement in BLEU scores and precision metrics, particularly when the sample size was increased to 1000 sentences. This larger sample size helped to better showcase the efficacy of the targeted interventions. Out of the evaluated sentences, 45 showed improved translation quality while 42 exhibited some regression. However, the magnitude and quality of improvements were substantially more significant than the regressions, underscoring the potential benefits of fine-tuning MT systems with carefully curated data.

\section{Discussion}
\subsection{Interpretation of Results}
Our experiments provide valuable insights into the role of closed-class words in machine translation, particularly from Chinese to English. Initial testing and fine-tuning of the Helsinki NLP model with a dataset specifically enhanced for attributive nouns showed nuanced but measurable improvements in translation accuracy. The slight increase in BLEU scores and precision metrics, especially noticeable after expanding the sample size, suggests that targeted interventions on specific linguistic features, such as attributive nouns, can significantly enhance machine translation system performance. This improvement is critical because attributive nouns in Chinese often omit particles like DE, leading to significant ambiguities when translated into English without proper contextual handling.

Our findings emphasize the importance of accurately translating closed-class words. These words, while not carrying separate meanings themselves, crucially structure the syntax and semantics of sentences. Incorrect translation or omission can disrupt the logical flow of the translated text, resulting in outputs that are grammatically correct but semantically flawed. Moreover, the improved handling of these words through manual adjustments and model tuning illustrates that even minor enhancements in linguistic structure treatment can substantially improve the clarity and coherence of translated texts. This is particularly vital for languages like Chinese, where the omission of specific function words is a common practice and poses significant challenges in automated translation contexts.

\subsection{Limitations and Future Research Directions}
While the improvements are encouraging, the relatively modest increases in BLEU scores highlight the limitations of current machine translation technologies in handling complex linguistic phenomena. The necessity for manual intervention in creating the tuning dataset also reveals a significant gap in automated systems' ability to comprehend and reproduce nuanced linguistic features independently. Initially, we attempted to use a script (using Python's Jieba package) to automatically identify nouns by tags and append the missing particle ``DE.'' However, due to the multifunctional nature of many Chinese words, which can serve as nouns, verbs, or adjectives, this script did not perform adequately, leading us to manually insert ``DE'' in the original text to ensure data accuracy. This manual process was time-consuming and significantly limited the corpus's scope, thus constraining our ability to produce and observe the impact of a larger dataset. Future research should explore the development of more sophisticated NLP models that incorporate deeper linguistic analyses into the translation process, potentially reducing the need for manual adjustments. Expanding the scope of the study to include other languages with similar linguistic features could further our understanding of the universal applicability of our approach. Additionally, investigating the use of advanced machine learning techniques such as deep learning and neural networks may provide new avenues to automate the identification and correct translation of closed-class words more effectively, potentially leading to significant advancements in the field, especially for languages with complex syntactical structures.

\section{Conclusion}
\label{sec:conclusion}
In conclusion, our study demonstrates the potential benefits of targeted linguistic interventions in machine translation. By focusing on the accurate translation of closed-class words, especially in languages like Chinese where these words play a crucial grammatical role, we can significantly enhance the semantic accuracy of translated texts. Continued exploration and refinement of these techniques are essential for advancing the capabilities of machine translation technologies in the coming years.

\begin{table*}[]
\centering
\begin{tabular}{l|cc|cc|cc}
\hline
\multirow{2}{*}{Model} & \multicolumn{2}{c|}{Pre-trained}                     & \multicolumn{2}{l|}{Fine-Tuned - 60 sent}            & \multicolumn{2}{l}{Fine-Tuned - 1k sent}            \\ \cline{2-7} 
                       & \multicolumn{1}{c}{BLEU} & \multicolumn{1}{c|}{CHRF} & \multicolumn{1}{c}{BLEU} & \multicolumn{1}{c|}{CHRF} & \multicolumn{1}{c}{BLEU} & \multicolumn{1}{c}{CHRF} \\ \hline
MarianNMT  &  37.8    & 70.6    & 37.8  & 70.6 & 38.2   & 72.1  \\
NLLB-200   &  38.3 & 72.1  & 38.3  &  72.2  & 39.5 & 75.1  \\
mBART  & 36.7  & 64.8   & 36.7  &  64.8  & 37.1  & 66.4   \\ \hline
\end{tabular}
\caption{BLEU and CHRF results of the 3 NMTs used to measure the performance of our test-dataset}
\label{tbl:results}
\end{table*}

\section*{Acknowledgments} We extend our gratitude to the institutions that provided invaluable support. Our affiliates include New York University, Northeastern University, Pace University, Carnegie Mellon University, Universitat Pompeu Fabra, and the Barcelona Computing Center. 

\bibliography{custom}



\end{document}